\documentclass[conference]{IEEEtran}
\usepackage[T1]{fontenc}
\usepackage[none]{hyphenat}
\usepackage{cite}
\usepackage{amsmath,amssymb,amsfonts}
\usepackage{algorithmic}
\usepackage{graphicx}
\usepackage{textcomp}
\usepackage[most]{tcolorbox}
\usepackage[export]{adjustbox}
\usepackage{subcaption}
\usepackage{array}
\usepackage{caption}
\usepackage{url}
\usepackage{xcolor}
\usepackage{multirow}

\usepackage{multirow}
\usepackage{booktabs}
\usepackage{eso-pic}
\AddToShipoutPictureBG*{%
  \AtPageUpperLeft{%
  \raisebox{-1cm}[0pt][0pt]{%
    \makebox[\paperwidth][c]{%
      \parbox{\textwidth}{\centering\color{gray!30}\small
        This work has been submitted to the IEEE for possible publication. Copyright may be transferred without notice, after \\
which this version may no longer be accessible
        }
        }
        }
  }%
}
\def\BibTeX{{\rm B\kern-.05em{\sc i\kern-.025em b}\kern-.08em
    T\kern-.1667em\lower.7ex\hbox{E}\kern-.125emX}}
\begin{document}

\title{Object Counting with GPT-4o and GPT-5: A Comparative Study}
\author{\IEEEauthorblockN{Richard F\"{u}zess\'{e}ry\IEEEauthorrefmark{1}, Kaziwa Saleh\IEEEauthorrefmark{2}\IEEEauthorrefmark{3}, S\'{a}ndor Sz\'{e}n\'{a}si\IEEEauthorrefmark{3}\IEEEauthorrefmark{4}, Zolt\'{a}n V\'{a}mossy\IEEEauthorrefmark{3}}
\IEEEauthorblockA{\IEEEauthorrefmark{1}Software Engineering Institute, Obuda University, Budapest, Hungary \\}  
\IEEEauthorblockA{\IEEEauthorrefmark{2}Doctoral School of Applied Informatics and Applied Mathematics, Obuda University, Budapest, Hungary \\}  
\IEEEauthorblockA{\IEEEauthorrefmark{3}John von Neumann Faculty of Informatics, Obuda University, Budapest, Hungary \\}
\IEEEauthorblockA{\IEEEauthorrefmark{4}Faculty of Economics and Informatics, J. Selye University, Komárno, Slovakia\\
Emails: richard.fuzessery@stud.uni-obuda.hu, \{kaziwa.saleh, szenasi.sandor, vamossy.zoltan\}@nik.uni-obuda.hu}

}
\maketitle

\begin{abstract}
Zero-shot object counting attempts to estimate the number of object instances belonging to novel categories that the vision model performing the counting has never encountered during training. Existing methods typically require large amount of annotated data and often require visual exemplars to guide the counting process. However, large language models (LLMs) are powerful tools with remarkable reasoning and data understanding abilities, which suggest the possibility of utilizing them for counting tasks without any supervision. In this work we aim to leverage the visual capabilities of two multi-modal LLMs, GPT-4o and GPT-5, to perform object counting in a zero-shot manner using only textual prompts. We evaluate both models on the FSC-147 and CARPK datasets and provide a comparative analysis. Our findings show that the models achieve performance comparable to the state-of-the-art zero-shot approaches on FSC-147, in some cases, even surpass them.
\end{abstract}

\begin{IEEEkeywords}
Zero-shot counting, large language models, LLMs.
\end{IEEEkeywords}

\section{Introduction}
Counting is a fundamental visual skill that humans perform effortlessly across nearly all object types, from animals in a herd to fruits in a bowl, without explicit supervision. This innate capability relies on general visual understanding rather than category-specific learning. In contrast, modern computer vision systems struggle to replicate this flexibility. Most counting models are designed to estimate counts for a single object class such as animals \cite{arteta2016counting}, crowd \cite{wang2020distribution, ma2019bayesian, zhang2016single}, vehicles \cite{hsieh2017drone}, or microscopic cells \cite{xie2015cell}, with limited ability to generalize beyond their training domain. Therefore, recent works focus on open-world object counting, which unlike conventional counting approaches bound to pre-defined categories, are designed to enumerate anything including objects they have not seen before. 

Existing methods require training on enormous quantity of annotated data and precise dot-level labels, and acquiring such annotations is an expensive and time-consuming task. Most of the available datasets \cite{hsieh2017drone, idrees2013multi, idrees2018composition, sindagi2020jhu, wang2020nwpu, zhang2016single} are typically restricted to individual object categories, which prevent models from learning general visual counting ability and limit their capacity to generalize to unseen categories. To overcome these problems, few-shot counting can reduce the annotation cost by enabling models to adapt to novel categories using only a few labeled examples. In this work, however, we attempt to address the previously mentioned problems with a zero-shot approach. 

Recent LLMs cannot only generate text, but their capability has evolved far beyond this, they have gained a remarkable capability in visual reasoning and contextual understanding. These abilities can be harnessed to interpret visual content, therefore, they are increasingly being integrated into vision tasks such as object detection, image captioning, and instance segmentation \cite{wang2023visionllm}. Especially, the emergence of multi-modal architectures like GPT-4 \cite{achiam2023gpt} and GPT-5 have further accelerated this fusion, as they are being applied for visual reasoning tasks \cite{ray2023chatgpt}, such as medical reasoning \cite{safari2025performance}, order recovery \cite{saleh2025gpt}, amodal completion \cite{fan2025multi}, occluded object counting \cite{pothiraj2025capture}, and occluded object understanding \cite{qiu2024occ}.

In this work, we use pre-trained GPT-4o and GPT-5 models to count objects in an image in zero-shot manner. The models are given the image and a prompt, in which we specify the category of the object(s) in question and the models will generate a numerical count. 

We tested the models on FSC-147 \cite{ranjan2021learning} and CARPK \cite{hsieh2017drone} datasets and the results demonstrate that both models can provide comparable and in some cases better counting results compared to the baselines, even if they are not specifically designed for such a task in zero-shot manner.

In summary, our contributions are as follows: First, we address the challenge of open-world object counting by leveraging two versions of GPT, effectively removing the need for training on large-scale counting datasets. Second, we investigate the counting capabilities of GPT-4o and GPT-5, using text-only prompts. Third, through a comparative evaluation on two established counting benchmarks, FSC-147 and CARPK, we show that GPT-5 consistently delivers higher accuracy and more reliable predictions than GPT-4o on FSC-147, while both models show weak performance on CARPK compared to the baseline models.

\section{Related Work}
\subsection{Few-Shot and Zero-Shot Counting}
Few-Shot counting reduce the annotation cost by training models to adapt to novel categories with few labeled data. Liu et al. \cite{Liu_2022_BMVC} utilize transformers and arbitrary number of exemplars in CounTR for open-world counting. Similarly, Duki\'c et al. \cite{djukic2023low} employ LOCA to iteratively merge the shape and appearance features of the exemplars with the image features. On the other hand, FamNet \cite{ranjan2021learning} combines a multi-scale feature extraction module and a class-agnostic density prediction module to handle objects from novel categories. Vision Transformers (ViT) are used in CACViT \cite{wang2024vision} that applies self-attention to perform feature extraction and similarity matching simultaneously. 
Amini-Naieni et al. \cite{amini2024countgd} propose CountGD that leverages the capability of GroundingDINO \cite{shilong2025grounding} to improve the capability of the model to generalize across categories and increase the counting accuracy.  

On the other hand, utilizing text prompts in place of exemplars, Zero-shot Object Counting (ZSC) \cite{xu2023zero} enables specifying objects without requiring any training data for the new target categories. Several works employ pre-trained vision language models (VLMs) such as CLIP \cite{radford2021learninga}. For example, CLIPCount \cite{jiang2023clip} applies CLIP to align text embedding with patch-level visual features. Also Kang et al. utilize CLIP in VLCount \cite{kang2024vlcounter} to improve text-image alignment. Other researchers \cite{zhu2025vacount} have employed GroundingDINO to select exemplars that each contains exactly one instance between the bounding box candidates. SAM \cite{kirillov2023segment} is used in PseCo \cite{huang2024point} to generate mask proposals for all potential objects, while CLIP classifies these proposals to produce precise object counts. In SAVE \cite{zgaren2025save}, authors employ YOLOv8 \cite{yolov8_ultralytics} to extract image feature maps while preserving spatial properties and then applies a contextual transformer for semantic embedding. 

\subsection{GPT in Vision}
Recently, several works have utilized the capability of GPT-4o to interpret and understand an image. For example, Fan et al. \cite{fan2025multi} employ GPT-4o to reason about occlusion relationships and perform boundary expansion on occluded object to achieve amodal completion. Additionally, Saleh et al. \cite{saleh2025gpt} have applied GPT-4 for occlusion order recovery. 
In case of counting task, authors in \cite{pothiraj2025capture} evaluate the performance of several VLMs including GPT-4o on a benchmark that evaluates spatial reasoning abilities under occlusion called CAPTURe. Similarly, Qharabagh et al. \cite{qharabagh2024lvlm} test the counting capability of GPT-4o and several other VLMs. To enhance their performance on crowded images, they propose a method that partitions each image into smaller sub-images and counts the target objects within each patch.
However, our work differs from the aforementioned studies in that we rely solely on textual prompts without supplying additional information, and we process each image in full using both GPT-4o and GPT-5.

\section{Datasets}
\textbf{FSC-147:} FSC-147 \cite{ranjan2021learning} is a dataset for open-world object counting, specifically designed for few-shot counting tasks. It consists of 6,135 images divided into training, validation, and test sets, with no overlap in object classes between them. The training set includes 89 classes, while the validation and test sets each contain 29 classes. We use the object class of each image and embed it into the prompt to guide the models during the counting process.

It is worth mentioning that while running the models on the dataset, we found 2 image classes that were annotated incorrectly. The label says that the object which should have been counted was a cupcake or a donut tray. However, in the images, there was one tray of either classes, but the point annotations instead represented the count of the individual items on the trays, rather than the trays themselves.

\textbf{CARPK:} CARPK \cite{hsieh2017drone} is a dataset of cars in aerial images of parking lots captured by drones. It comprises 989 images in the training set and 459 images in the test set.

\setlength{\fboxsep}{0pt}
\setlength{\fboxrule}{0.8pt}
\begin{figure*}[!h]
\centering
    \begin{subfigure}[t]{0.2\textwidth}
        \fbox{\includegraphics[width=\textwidth,height=0.08\textheight]{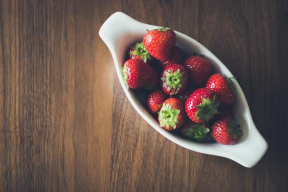}}
        \caption{Class: strawberries, GT: 12, P$_{GPT-4o}$: 10, P$_{GPT-5}$: 12}
    \end{subfigure}
    \hfill
        \begin{subfigure}[t]{0.2\textwidth}
        \fbox{\includegraphics[width=\textwidth,height=0.08\textheight]{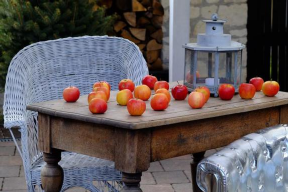}}
        \caption{Class: apples, GT: 21, P$_{GPT-4o}$: 20, P$_{GPT-5}$: 20}
    \end{subfigure}
    \hfill
    \begin{subfigure}[t]{0.2\textwidth}
        \fbox{\includegraphics[width=\textwidth,height=0.08\textheight]{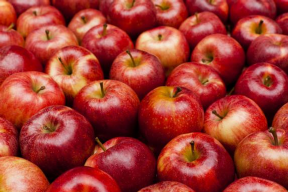}}
        \caption{Class: apples, GT: 35, P$_{GPT-4o}$: 33, P$_{GPT-5}$: 35}
    \end{subfigure}
    \hfill
    \begin{subfigure}[t]{0.2\textwidth}
        \fbox{\includegraphics[width=\textwidth,height=0.08\textheight]{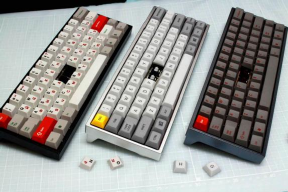}}
        \caption{Class: keyboard keys, GT: 184, P$_{GPT-4o}$: 210, P$_{GPT-5}$: 207}
    \end{subfigure}
    
    \vspace{1em}
    
    \begin{subfigure}[t]{0.2\textwidth}
        \fbox{\includegraphics[width=\textwidth,height=0.08\textheight]{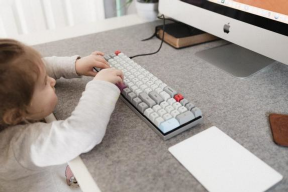}}
        \caption{Class: keyboard keys, GT: 70, P$_{GPT-4o}$: 61, P$_{GPT-5}$: 66}
    \end{subfigure}
    \hfill
     \begin{subfigure}[t]{0.2\textwidth}
        \fbox{\includegraphics[width=\textwidth,height=0.08\textheight]{images/1509.pdf}}
        \caption{Class: keyboard keys, GT: 50, P$_{GPT-4o}$: 52, P$_{GPT-5}$: 37}
    \end{subfigure}
    \hfill
    \begin{subfigure}[t]{0.2\textwidth}
        \fbox{\includegraphics[width=\textwidth,height=0.08\textheight]{images/2005.pdf}}
        \caption{Class: sunglasses, GT: 140, P$_{GPT-4o}$: 132, P$_{GPT-5}$: 144}
    \end{subfigure}
    \hfill
    \begin{subfigure}[t]{0.2\textwidth}
        \fbox{\includegraphics[width=\textwidth,height=0.08\textheight]{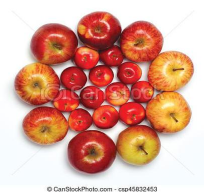}}
        \caption{Class: apples, GT: 21, P$_{GPT-4o}$: 21, P$_{GPT-5}$: 23}
    \end{subfigure}
\caption{Examples of results produced by GPT-4o and GPT-5 using the second prompt. The predicted counts from each model are shown as P$_{GPT-4o}$ and P$_{GPT-5}$, respectively. In most cases, GPT-5 generates more accurate predictions, however, GPT-4 sometimes performs better.}
\label{fig:results}
\end{figure*}

\section{Method}
We leverage the counting capabilities of the pre-trained GPT-4o and GPT-5 models, evaluating them on both the validation and test sets of the FSC-147 dataset, as well as on the test set of the CARPK dataset. For each image, the corresponding object class is embedded into a text prompt, which is then provided to the models to obtain a numerical count. Since the phrasing of the prompt influences model performance, we designed multiple prompts with varying levels of complexity (see Fig.~\ref{fig:prompt}). Each prompt was initially tested with a single run to measure error rates, followed by five repeated runs whose results were averaged for stability. For the CARPK dataset, which contains only one object category, all prompts used the class label ``car''.

\begin{figure}[ht]
\centering
\begin{tcolorbox}[colback=gray!5, colframe=black,]
\textbf{Prompt \#1:}\textit{ Count all distinct instances of bird visible in the image, including those that are partially visible, slightly obscured, blurred, or touching others, as long as there is reasonable visual evidence to identify them as individual chopsticks. In clustered or overlapping areas, estimate the number of individual chopsticks present. Return only the final count as a plain integer with no additional text.}
\end{tcolorbox}
\begin{tcolorbox}[colback=gray!5, colframe=black,]
\textbf{Prompt \#2:}\textit{ Look at the image and identify only birds. Count each birds, including partially visible ones, and avoid double-counting. Return only the final count as a plain integer with no additional text.}
\end{tcolorbox}
\caption{Example of prompts provided to the pre-trained models for object counting. The object classes were obtained directly from the dataset.}
\label{fig:prompt}
\end{figure}

The models are evaluated on FSC-147 and CARPK datasets. The performance of the models are assessed through Mean Absolute Error (MAE) and Root Mean Squared Error (RMSE). For an image $X_{i}$, MAE and RMSE are defined as:

\begin{equation}
\text{MAE} = \frac{1}{N} \sum^{N}_{i = 1} | \hat{c}_i - c_{i}|, \quad \text{RMSE} = \sqrt{ \frac{1}{N} \sum^{N}_{i = 1} (\hat{c}_i - c_{i})^2}
\end{equation}

where $N$ is the number of test images. $c_{i}$ and $\hat{c}_i$ are the ground truth count and predicted count by the model, respectively. 

\section{Results and Discussion}
The results indicate that GPT-4o and GPT-5 achieve either the best or second-best performance on the FSC-147 test set when using the second prompt, while the first prompt consistently produces lower error rates (see Table~\ref{table:results}). Between the two, GPT-5 demonstrates greater accuracy in count estimation compared to GPT-4o. Although the models are not specifically designed for object counting, their performance is comparable to state-of-the-art zero-shot open-world counting models.

For the CARPK dataset, both models showed higher error rates across the two prompts, with GPT-5 performing slightly better than GPT-4o when given the first prompt. 

\begin{figure*}[!h]
\centering
    \begin{subfigure}[t]{0.2\textwidth}
        \fbox{\includegraphics[width=\textwidth,height=0.08\textheight]{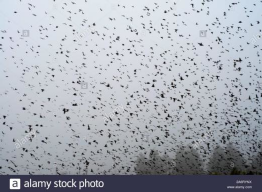}}
        \caption{Class: birds, GT: 2092, P$_{GPT-4o}$: 500, P$_{GPT-5}$: 980}
    \end{subfigure}
    \hfill
        \begin{subfigure}[t]{0.2\textwidth}
        \fbox{\includegraphics[width=\textwidth,height=0.08\textheight]{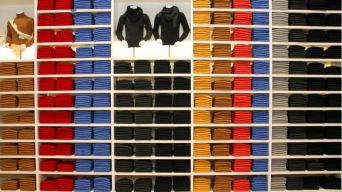}}
        \caption{Class: shirts, GT: 1231, P$_{GPT-4o}$: 116, P$_{GPT-5}$: 4}
    \end{subfigure}
    \hfill
    \begin{subfigure}[t]{0.2\textwidth}
        \fbox{\includegraphics[width=\textwidth,height=0.08\textheight]{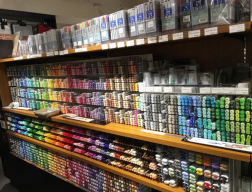}}
        \caption{Class: markers, GT: 3701, P$_{GPT-4o}$: 780, P$_{GPT-5}$: 620}
    \end{subfigure}
    \hfill
    \begin{subfigure}[t]{0.2\textwidth}
        \fbox{\includegraphics[width=\textwidth,height=0.08\textheight]{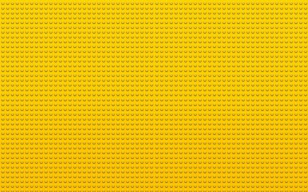}}
        \caption{Class: legos, GT: 2560, P$_{GPT-4o}$: 1, P$_{GPT-5}$: 1}
    \end{subfigure}
\caption{Example of images where the pre-trained GPT-4o and GPT-5 models do not produce correct counting predictions. P$_{GPT-4o}$, P$_{GPT-5}$ present the predictions of both models.}
\label{fig:failureSamples}
\end{figure*}

However, both models produce significant errors when processing highly crowded images in the FSC-147 dataset. Figure~\ref{fig:failureSamples} illustrate several examples along with ground truth and predicted counts. For instance, in subfigure (d), which contains 2,500 lego pieces, the models incorrectly perceive the entire collection as a single object and predict a count of 1. Table~\ref{table:failureImages} lists the validation and test images where both models fail. The first prompt produces incorrect predictions for more images compared to the second prompt.

In addition to the previously mentioned prompts, we experimented with several others varying in complexity and detail. However, none consistently achieved higher accuracy than the two selected prompts; therefore, they were excluded from further analysis. However, we observed that prompts that are either too long or too short result in less accurate predictions.

\begin{table}[!h]
\caption{Comparison results of pre-trained GPT-4o and GPT-5 against the state-of-the-art methods for zero-shot open-world object counting. Models with * are the ones where the images with high error rate are skipped. All results are for text-based prompts, except for RichCount $\dagger$ which uses a description. The best results are shown in bold, while the second-best results are underlined.}
\label{table:results}
\centering
    \begin{tabular}{  p{0.28\linewidth} | c c | c c } \hline  
    \multirow{2}{*}{Method} & \multicolumn{2}{c}{\textbf{Validation}} & \multicolumn{2}{c}{\textbf{Test}} \\ 
     & MAE $\downarrow$ & RMSE $\uparrow$ & MAE $\downarrow$ & RMSE $\uparrow$ \\ \hline
     \multicolumn{2}{l}{\textbf{FSC-147}}\\ \hline
     ZSC\cite{xu2023zero} & 26.93 & 88.63 & 22.09 & 115.17 \\
     VA-Count\cite{zhu2024zero} & \underline{17.87} & 73.22 & 17.88 & 129.31 \\
     CLIP-count\cite{jiang2023clip} & 18.79 & 61.18 & 17.78 & 106.62 \\
     VLCounter\cite{kang2024vlcounter} & 18.06 & 65.13 & 17.05 & 106.16 \\
     CountTX\cite{amini2023open} & \textbf{17.10} & 65.61 & \textbf{15.69} & 106.06 \\
     RichCount\cite{zhu2025expanding} & 18.65 & 58.55 & 16.37 & 102.48 \\ 
     GPT-4o prompt \#1 & 35.14 & 328.67 & 20.73 & 103.65 \\
     GPT-4o prompt \#$1^{*}$ & 20.32 & \underline{56.77} & 16.84 & 41.58\\
     GPT-4o prompt \#2 & 24.16 & 84.62 & 22.1 & 135.88 \\
     GPT-4o prompt \#$2^{*}$ & 22.0 & 63.44 & 16.87 & \textbf{38.34} \\
     GPT-5 prompt \#1 & 21.82 & 78.34 & 20.96 & 123.33 \\
     GPT-5 prompt \#$1^{*}$ & 19.03 & \textbf{51.77 }& \underline{16.24} & 41.63 \\
     GPT-5 prompt \#2 & 23.52 & 83.4 & 21.59 & 117.28 \\
     GPT-5 prompt \#$2^{*}$ & 21.27 & 59.52 & 17.12 & \underline{40.77} \\   
    \hline \hline 
     \multicolumn{2}{l}{\textbf{CARPK}}\\ \hline
     CLIP-count\cite{jiang2023clip} & - & - & \underline{11.96} & \underline{16.61} \\
     RCC\cite{hobley2022learning} & - & - & \textbf{8.13} & \textbf{10.87} \\
     GPT-4o prompt \#1 & - & - & 31.08 & 41.76 \\
     GPT-4o prompt \#2 & - & - & 38.57 & 52.61 \\
     GPT-5 prompt \#1 & - & - & 30.13 & 39.64 \\
     GPT-5 prompt \#2 & - & - & 40.41 & 51.9 \\
    \hline \hline
    \end{tabular}
\end{table}

\begin{table}[!h]
\caption{Failure cases where both GPT-4o and GPT-5 produce extremely large errors.}
\label{table:failureImages}
\centering
\resizebox{0.5\textwidth}{!}{
    \begin{tabular}{  p{0.3\linewidth} | c c c c } \hline  
    \multicolumn{5}{c}{\textbf{GPT-4o}}\rule{0pt}{10pt}\\ \hline
     Prompt & Image & Class & GT value & Model value\\ \hline
     \multirow{2}{*}{Prompt \#1 (Validation)} & 935.jpg & Birds & 2092 & 500 \\
      & 949.jpg & Birds & 1092 & 5000 \\
      & 3477.jpg & Polka dots & 344 & 2300 \\
      & 3665.jpg & Toilet paper rolls & 907 & 11481 \\
      & 7656.jpg & Shirts & 1231 & 116 \\
      \hline
     \multirow{2}{*}{Prompt \#1 (Test)} & 1123.jpg & Markers & 3701 & 780 \\ 
      & 7611.jpg & Legos & 2560 & 1 \\ 
      \hline
      \multirow{2}{*}{Prompt \#2 (Validation)} & 935.jpg & Birds & 2092 & 500 \\
      & 7656.jpg & Shirts & 1231 & 4 \\
      \hline
     \multirow{2}{*}{Prompt \#2 (Test)} & 1123.jpg & Markers & 3701 & 3 \\ 
      & 7611.jpg & Legos & 2560 & 1 \\
    \hline 
     \multicolumn{5}{c}{\textbf{GPT-5}}\rule{0pt}{10pt}\\ \hline
     Prompt & Image & Class & GT value & Model value\\ \hline
     \multirow{2}{*}{Prompt \#1 (Validation)} & 935.jpg & Birds & 2092 & 980 \\
      & 949.jpg & Birds & 1092 & 2400 \\
      & 7656.jpg & Shirts & 1231 & 4 \\
      \hline
     \multirow{2}{*}{Prompt \#1 (Test)} & 1123.jpg & Markers & 3701 & 620 \\ 
      & 7611.jpg & Legos & 2560 & 1 \\
      \hline
      \multirow{2}{*}{Prompt \#2 (Validation)} & 935.jpg & Birds & 2092 & 392 \\
      & 7656.jpg & Shirts & 1231 & 4 \\
      \hline
     \multirow{2}{*}{Prompt \#2 (Test)} & 1123.jpg & Markers & 3701 & 900 \\ 
      & 7611.jpg & Legos & 2560 & 1 \\ 
    \hline \hline
    \end{tabular}
    }
\end{table}

\section{Limitations} 
While we demonstrate the potential of leveraging the capability of multi-modal LLMs for open-world object counting, we address several limitations with our work. First, the models are not free to use, making large-scale evaluations financially demanding, especially GPT-5. Second, our method does not incorporate visual exemplars, relying solely on textual prompts to guide the counting process. This may reduce the ability of the models to differentiate between visually similar objects or avoid double counting. Finally, we embed only the object class name into the prompt without providing any additional details about the object's physical visual characteristics, which may reduce the interpretation accuracy of the models.

\section{Conclusions}
Object counting plays a curial role in a wide range of domains, including traffic monitoring and analysis, manufacturing and logistics, agriculture, environmental monitoring, security, and healthcare. Therefore, accurate estimation of object quantities is essential. In this work, we explore the potential of leveraging the visual reasoning and understanding capabilities of two multi-modal LLMs, GPT-4o and GPT-5, for object counting in a zero-shot manner. Our approach relies solely on textual prompts to retrieve the counting result. Extensive results demonstrate the effectiveness of both models on object counting. Our findings present that GPT-5 can, in most cases, achieve more accurate predictions compared to GPT-4o. In the future, the models can be experimented with descriptions derived from VLMs, such as CLIP, to enable more precise and context-aware object counting.

\section*{Acknowledgment}
We would like to thank the ``Doctoral School of Applied Informatics and Applied Mathematics'' at Obuda University, and the National Talent Program (NTP--HHTDK--25-0026) for their invaluable support. We also acknowledge the ``Examining occlusions using deep machine learning'' project for providing access to the HUN-REN Cloud \cite{heder2022past} which helped us in obtaining the results presented in this paper.

\bibliographystyle{ieeetr}
\bibliography{references}

\end{document}